\documentclass[pdflatex,sn-mathphys-num]{sn-jnl}
\usepackage{graphicx}%
\usepackage{multirow}%
\usepackage{amsmath,amssymb,amsfonts}%
\usepackage{amsthm}%
\usepackage{mathrsfs}%
\usepackage[title]{appendix}%
\usepackage{xcolor}%
\usepackage{textcomp}%
\usepackage{manyfoot}%
\usepackage{booktabs}%
\usepackage{algorithm}%
\usepackage{algorithmicx}%
\usepackage{algpseudocode}%
\usepackage{listings}%
\usepackage{graphicx}
\usepackage{subcaption}
\usepackage{caption}
\usepackage{url}
\usepackage{hyperref}

\begin{document}

\title[Article Title]{EmoACT: a Framework to Embed Emotions into Artificial Agents Based on Affect Control Theory}

\author*[1]{\fnm{Francesca} \sur{Corrao}}\email{francesca.corrao@edu.unige.it}

\author[1]{\fnm{Alice} \sur{Nardelli}}\email{alice.nardelli@edu.unige.it}

\author[2]{\fnm{Jennifer} \sur{Renoux}}\email{Jennifer.Renoux@oru.se}

\author[1]{\fnm{Carmine Tommaso} \sur{Recchiuto}}\email{carmine.recchiuto@dibris.unige.it}

\affil*[1]{\orgdiv{Department of Informatics, Bioengineering, Robotics and System Engineering}, \orgname{Università degli studi di Genova}, \orgaddress{\street{Via all'Opera Pia 13}, \city{Genova}, \postcode{16145}, \country{Italy}}}

\affil[2]{\orgdiv{Center for Applied Autonomous Sensor Systems}, \orgname{Örebro University}, \orgaddress{ \city{Örebro}, \postcode{70182}, \country{Sweden}}}

\abstract{
As robots and artificial agents become increasingly integrated into daily life, enhancing their ability to interact with humans is essential.
Emotions, which play a crucial role in human interactions, can improve the naturalness and transparency of human-robot interactions (HRI) when embodied in artificial agents.
This study aims to employ Affect Control Theory (ACT), a psychological model of emotions deeply rooted in interaction, for the generation of synthetic emotions.
A platform-agnostic framework inspired by ACT was developed and implemented in a humanoid robot to assess its impact on human perception.
Results show that the frequency of emotional displays impacts how users perceive the robot.
Moreover, appropriate emotional expressions seem to enhance the robot's perceived emotional and cognitive agency.
The findings suggest that ACT can be successfully employed to embed synthetic emotions into robots, resulting in effective human-robot interactions, where the robot is perceived more as a social agent than merely a machine.
}

\keywords{Synthetic Emotions, Human-Robot Interaction, Social Robotics, Affect Control Theory}

\maketitle

\section{Introduction}\label{sec1}
The interest in social robots continues to rise due to their capability to assist people in multiple areas, from assistive tasks~\cite{chu_service_2017} and teaching scenarios~\cite{woo_use_2021} to simply serving as companions in everyday life~\cite{zsiga_evaluation_2018}.
The goal of these robots is to interact naturally with humans.
Previous studies have shown that their performance, in terms of acceptability, can be improved if they are capable of perceiving and displaying emotions~\cite{stock-homburg_survey_2022}.
Indeed, emotions are not merely correlated with an internal state but are deeply rooted in human interactions and play a fundamental role in mitigating bad news, indicating common ground, and more~\cite{gendron_reconstructing_2009}.
Integrating emotions into robots enhances the transparency of interactions~\cite{stock-homburg_survey_2022}, allowing humans to better understand and predict the robot's behavior.
Furthermore, the presence of emotions enables humans to rely on the same cues they use to communicate with each other when interacting with robots.
As a result, the acceptability of emotional robots increases, and the bond between humans and robots strengthens~\cite{deukey_lee_general_2009}.
Effectively embedding emotions into artificial agents can reduce unfamiliarity with artificial systems by making their behavior more predictable, understandable, and transparent~\cite{bethel_survey_2008}.

In order to achieve the full integration of synthetic emotions in artificial agents, three aspects need to be considered: their modeling within the system, their generation process, and how they are displayed.
The goal in all three areas is to replicate and adapt human behavior.
Therefore, a preliminary study of human emotions is fundamental.
One of the most widely applied psychological theories is Cognitive Appraisal Theory \cite{moors_appraisal_2013}, which was employed in one of the pioneering projects on synthetic emotions \cite{breazeal_emotion_2003}.
This theory continues to be used in subsequent work, applied in different scenarios and incorporating aspects such as the modeling of ongoing relationships and personalities~\cite{kowalczuk_computational_2016}.
Other studies have explored alternative approaches for generating emotions, such as Markov Chains \cite{ficocelli_promoting_2016}, but no psychological theory other than Cognitive Appraisal has been extensively investigated in this process.
However, psychology research on emotions presents multiple theories beyond Cognitive Appraisal \cite{gendron_reconstructing_2009}, including Affect Control Theory (ACT)~\cite{stets_emotions_2014}.
According to ACT, emotions help individuals perceive social relationships, regulate their evolution, and communicate their nature~\cite{stets_emotions_2014}.
ACT posits that emotions arise during interactions from the discrepancy between an individual's identity and the impression they believe others have of them.
This process is described by equations in a three-dimensional space called EPA space, which measures three aspects of emotions, identity, and impression: Evaluation, Potency, and Activity.

Unlike Cognitive Appraisal Theory, where emotions are not directly tied to interactions but rather arise from a personal evaluation of events in pursuit of goals, Affect Control Theory (ACT) explicitly links emotions to social interactions.
This distinction is particularly relevant in the context of Human-Robot Interaction (HRI), where the goal is to create more natural interactions between humans and robots.
ACT’s focus on how emotions emerge from interactions makes it highly relevant for the generation of synthetic emotions in artificial agents.
This, combined with the lack of theoretical and applied research employing ACT in artificial agents, captured our interest and led us to the following research question:

\begin{center}
\textbf{RQ}: Can emotions be embodied in artificial agents using Affect Control Theory?
\end{center}

To address this question, we developed an application-independent framework based on ACT's emotional equations.
Additionally, the framework generates cues to portray five basic emotions—Anger, Fear, Happiness, Neutral, and Sadness—according to the agent's capabilities.  

Two within-subject studies were conducted to evaluate the performance of the architecture.
The framework was implemented on a humanoid robot in a collaborative storytelling scenario, which was tested twice to assess the impact of emotional display frequency on the perception of the robot. Specifically:

\begin{itemize}
\item \textbf{Experiment 1}: We compared a robot that does not display any emotions with a robot, integrated into our framework, that displays emotions at a low frequency.
\item \textbf{Experiment 2}: We compared a robot that does not display any emotions with a robot, integrated into our framework, that displays emotions at a high frequency.
\end{itemize}

The results showed that an ACT-based emotional framework can successfully convey emotions using a humanoid robot.
Additionally, we found that the frequency of emotional displays affects the perception of the robot’s emotional capabilities.
Overall, we suggest the potential of using ACT to embed synthetic emotions into artificial agents, emphasizing the importance of emotional cues in shaping the perception of cognitive abilities.

The article is structured as follows: Section~\ref{soa} presents a review of previous works on synthetic emotions and ACT. Section~\ref{methodology} describes the proposed architecture, while Section~\ref{implementation} details its robotic implementation. Sections~\ref{evaluation} and \ref{results} outline the experimental setup designed to evaluate the framework and present the results, respectively. Finally, Sections~\ref{discussion} and \ref{conclusion} discuss the findings and draw conclusions.

\section{Background}
\label{soa}
The idea of equipping robots and artificial agents with synthetic emotions has been around for more than 20 years.
The growing interest in synthetic emotions stems from the crucial role emotions play in human interactions; their absence would deprive robots of an important communication cue \cite{osuna_development_2020}.
According to \cite{stock-homburg_survey_2022}, a robot’s social presence is closely tied to its ability to express emotions.
Moreover, emotional expression enhances the transparency and perception of human-robot interaction (HRI) \cite{bethel_survey_2008}.
These aspects make emotions a vital tool for improving interactions between social robots and humans, fostering better integration and acceptance \cite{deukey_lee_general_2009}.

Research on synthetic emotions focuses on three main areas: generation, modeling, and portrayal~\cite{kowalczuk_computational_2016}.
For generation, various components are developed to collect external stimuli and process them to determine the emotions the robot should express.
For modeling, integrating emotions into artificial agents requires defining how they will be represented within the system.
Finally, for portrayal, agents must communicate and display their synthetic emotional state using different approaches depending on their specific capabilities. Ultimately, the goal is to replicate and adapt human behaviors by leveraging insights from psychological studies on human emotions

Research in psychology presents multiple theories and approaches that explain how emotions function in humans.
These can be divided into four macro areas~\cite{gendron_reconstructing_2009}:

\begin{itemize}
\item The Cognitive Appraisal Theory of emotion suggests that emotions are an adaptive response that reflect the appraisal of features of the environment significant for the organism's well-being~\cite{moors_appraisal_2013}.

\item The Constructivism Theory is based on the idea that humans create a social reality out of the things they teach one another and apply to physical instances. This is applied to emotions, so physical changes (internal or external) become emotions when they are defined as such. These categories are learned through language, socialization, and other cultural artifacts~\cite{barrett_conceptual_2014}.

\item The Basic Approach to emotions is based on the idea that emotions are discrete, automatic responses to universally shared cultural and individual-specific events~\cite{ekman_what_2011}. They all serve a function that allows humans to adapt and evolve~\cite{tracy_evolutionary_2014}.

\item The Sociology Model of human emotions sees the construction of emotions as an interactive, ongoing process that unfolds within interactions and relationships, deriving their shape and meaning from a socio-cultural context~\cite{boiger_construction_2012}. An example is Affect Control Theory~\cite{stets_emotions_2014}.
\end{itemize}
 
These theories are not equally investigated when trying to embed emotions into artificial agents.
Concerning the generation of synthetic emotions, the most investigated theory has been Cognitive Appraisal Theory due to the correlation between one's emotions and the goal they are trying to reach~\cite{kowalczuk_computational_2016}.
Indeed, this theory was used in one of the pioneering studies on equipping social robots with emotions~\cite{breazeal_emotion_2003}, where the goal was to adapt Cognitive Appraisal Theory to engage people in natural and expressive face-to-face interactions.
Successive works on emotion generation have continued to focus mainly on Cognitive Appraisal Theory, employing it in general-purpose application chatbots~\cite{ehtesham-ul-haque_emobot_2024}, service robots~\cite{kwon_emotion_2007}, promoting multi-sensory integration~\cite{park_emotion_2007}, and more~\cite{kowalczuk_computational_2016}.
Some studies have been able to correlate the generation of emotions with synthetic personality by employing Cognitive Appraisal Theory~\cite{park_robots_2009} or using Fuzzy Kohonen Cluster Network (FKCN) \cite{han_robotic_2013}.
Other studies have tried to generate emotions without considering psychological models.
For example, previous work used Markov chains to determine the emotional state of a socially assistive robot \cite{ficocelli_promoting_2016} or derived emotions from environmental features using reinforcement learning based on visual cues: color, fractal, and face pareidolia effects~\cite{wong_robot_2013}.
Additionally, the relationship built through interaction between the robot and its user is important, and in~\cite{kinoshita_emotion-generation_2014}, this was considered to generate emotions by taking into account the degree of intimacy.
Lastly, one work proposes a theoretical framework to consider different theories of emotions~\cite{rosales_general_2019}.
It consists of a modular architecture where each component is constructed based on different psychological findings, enhancing their similarities and differences.
This theoretical architecture was used in the design phase of EmoACT to determine how to separate the different components, with the goal of incorporating additional psychological theories in the future.

Concerning emotional modeling, there are mainly two approaches used to represent emotions~\cite{cavallo_emotion_2018}:
\begin{itemize}
\item \textbf{Distinct emotions}: emotions are categorized by their associated labels.
This approach is based on Basic Emotion Theory and typically includes emotions such as Anger, Disgust, Fear, Happiness, Sadness, and Surprise~\cite{ekman_what_2011}.
This approach does not allow for mapping concepts such as the distance between emotions and makes it difficult to capture nuances within the same emotion.
\item \textbf{Emotional spaces}: emotions are mapped in either two- or three-dimensional spaces so they are described by a point within a space~\cite{hoffmann_mapping_2012}.
The most commonly used emotional space is the PAD (Pleasure - Arousal - Dominance) model~\cite{gillioz_mapping_2016}.
\end{itemize}

The portrayal of emotions is usually based on the association between emotional terms and human behaviors, sounds, and colors.
Indeed, the type of cues used to showcase emotional states depends on the type of agent that employs them.
They are mainly divided into two categories: verbal and non-verbal cues.

Verbal cues involve the strategic manipulation of tone, pitch, rhythm, and word emphasis~\cite{frick_communicating_1985, philippou-hubner_performance_2012, um_emotional_2020, galanis_investigating_1996}.
Speech generation also plays an important role, as the choice of words and sentence structure can convey specific emotional undertones.
This has been achieved by employing deep neural networks to express emotion modeled in the PAD space~\cite{zhang_emotional_2018}.

Non-verbal cues include facial expressions, posture, movement, color, and sound.
Facial expressions require agents with the capability to move facial features such as eyes, eyebrows, or mouths.
Emotional behavior is generated by explicitly specifying how these features need to change.
This can be done by building humanoid faces with different degrees of freedom~\cite{breazeal_emotion_2003, ahn_appropriate_2012, han_robotic_2013}, which are then actuated to mimic human muscle movements of faces.
An alternative is to project animated faces onto screens~\cite{kwon_emotion_2007, park_emotion_2007} or use LED matrices to display eyebrows and mouths~\cite{churamani_teaching_2017}.
Facial expressions can represent basic emotions with different levels of subtlety or even complex emotional expressions, as has been done in~\cite{deukey_lee_general_2009}.

Another field of study investigates conveying emotions through humanoid robots' body posture and movement.
In this case, research conducted on mannequins and actors~\cite{coulson_attributing_2004, wallbott_bodily_1998} is applied to humanoid robots, creating static body postures or movement animations to showcase distinct emotions such as Happiness, Anger, Sadness, and more~\cite{erden_emotional_2013}.
Some studies map posture to affective spaces, such as correlating head position with the AVS (Arousal, Valence, and Stance) affect space~\cite{beck_towards_2010}.
The possibility of correlating emotional behavior with a range of values in an emotional space could improve the expressive capability of agents, allowing them to showcase nuances of the same emotion according to its intensity.

Other non-verbal cues used to display emotions, especially in non-humanoid robots such as robotic arms or mobile robots, include color, sound, and movement.
Many studies associate colors with distinct emotions~\cite{steinvall_anders_colors_2007, takahashi_association_2018}, often based on how emotions are described linguistically~\cite{steinvall_anders_colors_2007}.
An example is the association of blue with sadness, as being sad is described metaphorically as “feeling blue” \cite{loffler_multimodal_2018}.
All these cues can be combined to enhance emotion recognition \cite{song_expressing_2017}.
However, care must be taken because multi-modality does not always improve the perception of emotions and may even obscure the emotional state~\cite{loffler_multimodal_2018}.

\subsection{Affect Control Theory}
Affect Control Theory is a cross-disciplinary collaborative theory of emotions that links individual and social aspects of emotions and describes how emotions are affected by social mechanisms such as interactions, relationships, and culture~\cite{stets_emotions_2014}.
ACT sees emotion arising from an automatic and unconscious comparison between one's impression of self and one's identity.
In ACT, identity, impression, and emotions are grounded in a three-dimensional affective space.
The three values measured are Evaluation, Potency, and Activity, on a scale that ranges from -4 to +4.
The discrepancy between one's identity and the impression others have of them generates emotions.
Interactions play a key role in how this discrepancy evolves; during interaction, the goal is to make the discrepancy disappear, particularly when it is negative.
Identity also holds an important role in ACT, as it sets the characteristic emotions one is expected to feel, incorporating different aspects such as fixed individual traits, situational identity, and transient mood.
The display of emotions is regulated by emotional norms, which define their idealized portrayal.
ACT emphasizes the impact of social categorization on emotions, which are shaped by factors such as gender, culture, and racial group.
Emotional norms emerge from culturally shared sentiments regarding behavior, identities, and settings, reflecting collective expectations regarding emotional expression.

\section{EmoACT Proposed Framework}
\label{methodology}
\begin{figure}[htb]
\centering
\includegraphics[width=0.76\linewidth]{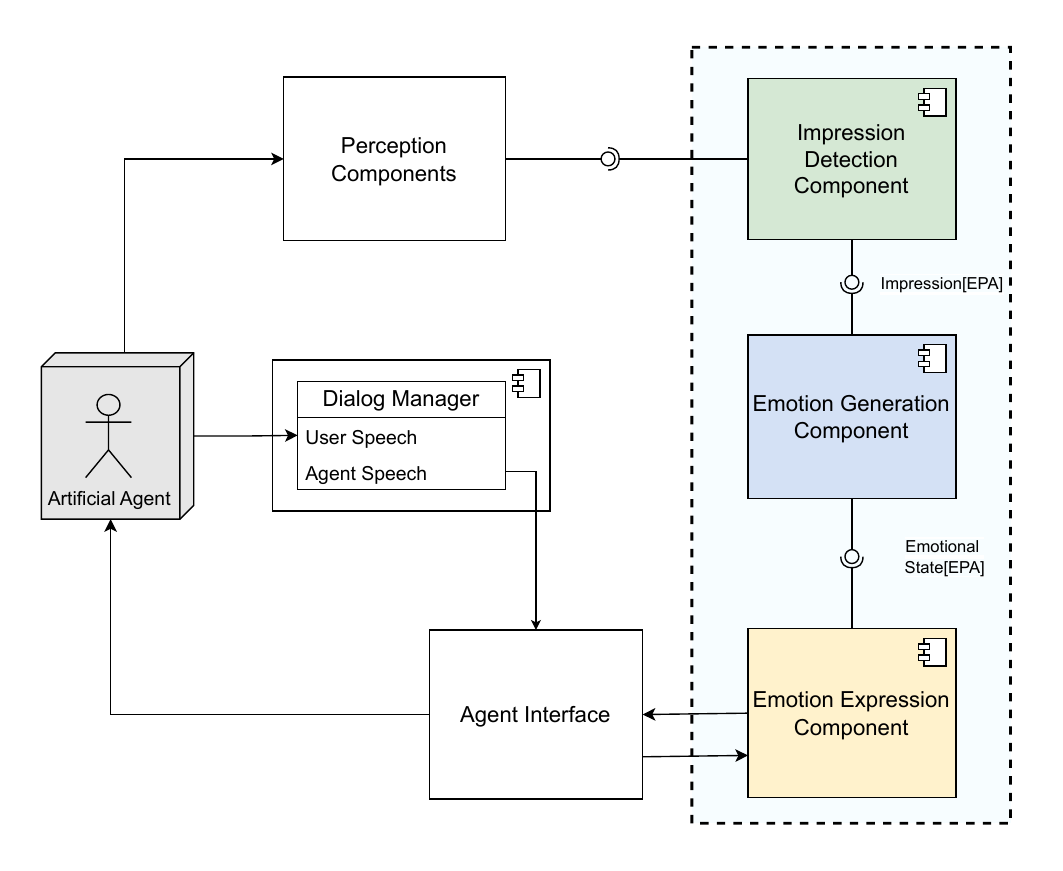}
\caption{EmoACT Framework.}
\label{fig:emoACT}
\end{figure}

This section introduces and describes the platform-agnostic components.
Unlike traditional approaches in the literature, the proposed solution is not dependent on a specific platform and can be applied across different artificial agents.
The objective of this work is to bridge the gap in the literature regarding the lack of ACT implementation by developing an application-independent framework that generates synthetic emotions based on ACT and provides the tools to showcase them.
According to ACT, individuals carry out two affective meanings during interaction: the fundamental and the transient ones.
The discrepancy between these two determines the emotion that is felt.
The proposed framework, shown in Figure~\ref{fig:emoACT}, tries to reproduce this behavior in an architecture composed of multiple interconnected modules.
The key components in this process are Impression Detection and Emotion Generation, which function as application servers.

The design of emoACT is inspired by the theoretical general framework proposed by~\citet{rosales_general_2019} and adapted to ACT.
In this approach~\cite{rosales_general_2019}, three main components are needed to achieve a complete integration of emotions: Affect, Feeling, and Emotional Behavior.
The Affect component is oriented toward the evaluation of the environment based on perceptual stimuli.
In emoACT, this is represented by the Impression Detection component, which is the module in charge of processing perception to evaluate the interaction.
The Impression Detection component (Section~\ref{ID}) receives input from the perception system, which consists of information about the user's emotional state, task, and agent-related cues.
It processes this data to estimate the impression the user has of the robot during the interaction and sends it to the Emotion Generation module.

The Feeling component represents the internal emotional state; this corresponds to emoACT's Emotion Generation component.
The Emotion Generation component (Section~\ref{EG}) contains the fundamental affective meaning of the agent.
It receives the impression the user has of the agent during interaction and, based on the emotional equations described by ACT~\cite{stets_emotions_2014}, produces the emotions the robot should portray.

Lastly, the Emotional Behavior component expresses the feeling through expression cues; this is implemented by the Emotional Expression component (Section~\ref{EE}).
It controls the expression of the internal emotional state by producing emotional behavior cues according to the capabilities of the artificial agents.
At this stage, the emotion, which has so far been modeled in EPA space, is mapped to the corresponding basic label and used to select the associated emotional act.

An addition to the theoretical framework~\cite{rosales_general_2019} is the Dialogue Manager, which handles the speech-based interaction between the human and the robot.
We chose to have humans interact with the robot through dialogue so that the interaction would be more natural.

\subsection{Impression Detection} \label{ID}
The Impression Detection component is a server that estimates the impression the user has of the robot during the interaction.
The impression is represented by a vector describing the three values in EPA space.
To estimate the impression, three perception cues are used: the user’s emotion, task-related input, and agent-related input.
These cues are provided by the perception components, which have been developed for the robot used in the experiments but can be easily adapted to work with other artificial agents.

The user's emotion is used to update the Evaluation value of the impression in EPA space.
This update either increases or decreases the Evaluation value based on the sign of the user's current emotion and is proportional to the difference between the current and previous emotional states.

The task- and agent-related perception cues considered were the user's gaze and their proximity to the robot.

The user's gaze is correlated with the Potency aspect but also affects other components of the impression.
Regarding Potency, if the user is looking at the robot, it indicates that the robot is effective in maintaining the user's attention.
This suggests that the robot is fulfilling its role and keeping the user's interest.
Additionally, when the user is looking at the robot, the robot has a greater impact on the user's emotional state.
In such cases, the effects of other components are more likely to be attributed to the robot's behavior rather than to external factors, leading to adjustments in the impression.
For example, if the user appears sad, the Evaluation value of the robot's identity will decrease.
If the user is looking at the robot, this sadness is likely due to the robot's behavior, further decreasing the Evaluation value.
On the other hand, if the user is looking elsewhere, their sadness is more likely caused by external factors, resulting in an increase in the Evaluation value.

Lastly, proximity correlates with the Activity and Evaluation aspects, as found in the study by \citet{fauville_impression_2022}.
According to this study, individuals closer to the camera are perceived as good and active, while those farther away are seen as passive and bad.
This information has been adapted by increasing or decreasing the Evaluation and Activity of the impression proportionally to changes in proximity.

The server waits for new information from the perception nodes and, once it receives input, calls the corresponding methods to update the impression.
If the impression has been updated, it is fed into the Emotion Generation component.

\subsection{Emotion Generation} \label{EG}
The Emotion Generation component is a server that computes and stores the synthetic emotion the agent should portray.
The emotional state is updated each time a new impression is received from the Impression Detection node.
Emotions are generated in EPA space by applying the equation for each of the three dimensions, resulting in a vector.

ACT describes how each element of emotion in EPA space evolves based on the discrepancy between an individual's identity and their perceived impression.
To apply these equations, it is essential to know the agent's identity, which is defined and stored internally as a vector in EPA space.
The computation of emotions across the three dimensions is detailed below.

The following equations are based on the discussion presented in~\cite{stets_emotions_2014}, which highlights the relationship between impression, identity, and emotions, particularly how the difference between impression and identity generates emotion.

The Evaluation value of emotions is influenced by two aspects of an individual's affective meaning: Evaluation and Activity.
The impact of Evaluation is described by Equation~\ref{eq:emo-ee}.
The sign of an emotion's Evaluation is directly proportional to the valence of one's impression, while its intensity varies with the extremity of the impression adjusted by the fundamental meaning~\cite{stets_emotions_2014}.
Activity also influences Evaluation, as shown in Equation~\ref{eq:emo-e-complete}.
Quiet identities (individuals with a negative or small Activity) tend to experience emotions with greater Evaluation~\cite{stets_emotions_2014}.
Additionally, the discrepancy in Activity levels between identity and impression leads to increased positivity when individuals are perceived as more active and a decrease when perceived as less active.

\begin{equation}
    \label{eq:emo-ee}
    emotion[E]_e = impression[E] - identity[E] + 1
\end{equation}
\begin{equation}
    \label{eq:emo-e-complete}
    emotion [E] = emotion[E]_e +(impression[A] - identity[A])*\delta
\end{equation}

The Potency of emotions depends on the Potency of the impression, with the identity setting expectations about it.
This relationship is described in Equation~\ref{eq:emo-pp}.
Powerless individuals who are perceived as more powerful experience high-Potency emotions, while powerful individuals who are perceived as less potent encounter low-Potency emotions~\cite{stets_emotions_2014}.
Activity discrepancies also influence Potency, as shown in Equation~\ref{eq:emo-p-complete}.
Negative-Potency emotions arise when the Activity discrepancy makes the individual appear overly frenetic, whereas positive-Potency emotions are experienced when the individual is perceived as quieter.

\begin{equation}
    \label{eq:emo-pp}
    emotion[P]_p = impression[P]-identity[P]
\end{equation}
\begin{equation}
    \label{eq:emo-p-complete}
    emotion[P] = Emotion[P]_p -(impression[A] - identity[A])
\end{equation}

The Activity value of emotions is derived from the discrepancy between the Activity level of identity and impression.
Emotions with positive Activity result from being perceived as more active, while emotions with negative Activity arise from perceptions of being overly inactive.
The confirmation of one's Activity level leads to moderately active emotions.
This relationship is expressed in Equation~\ref{eq:emo-a}.

\begin{equation}
    \label{eq:emo-a}
    emotion[A]= impression[A] + identity[A]
\end{equation}

These equations are implemented within the Emotion Generation module.
The node waits for updated impressions, and once received, it produces the corresponding emotional state.
Emotion updates occur instantaneously and do not take the previous emotional state into account.
However, the process appears dynamic due to the continuous updates of the impression component based on perception cues.
As a result, the dynamics of impression updates are directly reflected in emotion updates.
The current emotional state is stored and can be retrieved through server requests.

\subsection{Emotion Expression} \label{EE}
The Emotion Expression component is responsible for generating emotional behavior cues according to the agent's capabilities.
The type of cues available depends on the specific robot using the framework.
The developed version includes tools to portray emotions with the \href{https://www.aldebaran.com/en/pepper}{Pepper Robot by Aldebaran}.
However, this component, along with the overall framework, can be easily adapted to other artificial agents.

The module requests the emotional state from the Emotion Generator module and converts it from EPA space into the corresponding emotional label.
This conversion is performed using the cosine similarity function, which compares the emotional state vector with the predefined vectors of basic emotions.
The basic emotions considered are four of the seven described by~\citet{ekman_what_2011}: Anger, Fear, Happiness, and Sadness.
The emotion with the highest similarity above 60\% is selected.
If no emotion can be selected, the Neutral emotional state is chosen.
The EPA space values of the emotion labels can be seen in Table~\ref{table:emomap}.

\begin{table}[htb]
\centering
\begin{tabular}{lc}
\toprule
Emotion label  & Emotion EPA vector \\ \midrule
Angry     &[ 1.95, 1.34, 1.78 ]     \\
Fear      & [ -2.04, -0.94, -0.70 ] \\
Happiness & [ 3.54, 2.53, 1.28 ]   \\
Sadness   & [ -2.52, -2.29, -2.21 ] \\
\bottomrule
\end{tabular}
\caption{Association between emotion labels and their value in EPA space.}
\label{table:emomap}
\end{table}

\section{Implementation} \label{implementation}
To test the performance of the emotional framework in interaction with humans, it was implemented on the \href{https://www.aldebaran.com/en/pepper}{Pepper Robot by Aldebaran}.
To ensure control over the conversation flow and guarantee that all participants were exposed to all emotional states, a collaborative storytelling scenario was designed.
At specific points, the robot asks participants to decide how the story should proceed.
Participants are offered two choices at each decision point, and they must select one.
These choices were designed to be correlated with different impressions of the robot, specifically in terms of its Activity, Potency, and Evaluation levels.

This required adapting the perception and expression components and defining the inputs needed for the Impression Detection component.

\subsection{Impression Detection}
To integrate storytelling into the framework, the Impression Detection module (Section \ref{ID}) was modified to incorporate the effect of the user's choice when updating the impression.
In particular, each decision point in the story was associated with an expected Evaluation, Potency, and Activity (EPA) sign for the impression (Appendix \ref{stories}).
For example, at some point in the story, the human and the robot encounter an individual speaking a different language, and the robot offers to go ahead and interact with them since it knows the language.
If the user decides to send the robot, this would positively influence the robot’s Evaluation and Activity impression scores.
Conversely, rejecting the robot’s help would yield a negative impression.

When a choice is made, if the current impression sign matches the expected sign from the choice, the impression is incremented.
Otherwise, the impression values are adjusted to reflect the opposite sign.

\subsection{Perception Components}
This node processes information from the robot's sensors to obtain the user's current emotional state, distance from the robot, and gaze.
The information about proximity and gaze is directly obtained from the robot's equipped perception module and sent to the Impression Detection.
However, the robot's capability to detect human emotional expressions is insufficient for accurate estimation.
Therefore, the image acquired by the robot's camera is processed using the \href{https://www.morphcast.com/sdk/}{MorphCast SDK}, which provides information about the human's emotional state and attention level.

\subsection{Emotion Expression}
To showcase the emotional state using the Pepper robot, two types of emotional cues are used: whole-body animations and eye color.
These choices were made based on the robot's capabilities, which include performing smooth movements and changing the color of its eye LEDs.
The node directly connects to the robot and sends the emotional cues.

The eye color for each emotion is determined based on existing studies that explore the relationship between emotions and color~\cite{takahashi_association_2018, loffler_multimodal_2018}.
The associations used are shown in Table~\ref{table:emocolours}.

\begin{table}[htb]
\centering
\begin{tabular}{lc}
\toprule
Emotion label  & Eye color\\ \midrule
Angry          & Red\\
Fear           & Black\\
Happiness      & Green\\
Sadness        & Dark Blue\\
\bottomrule
\end{tabular}
\caption{Association between emotion labels and colors.}
\label{table:emocolours}
\end{table}

\begin{figure}[htb]
    \centering
    \begin{subfigure}[b]{0.23\textwidth}
    \includegraphics[width=\textwidth]{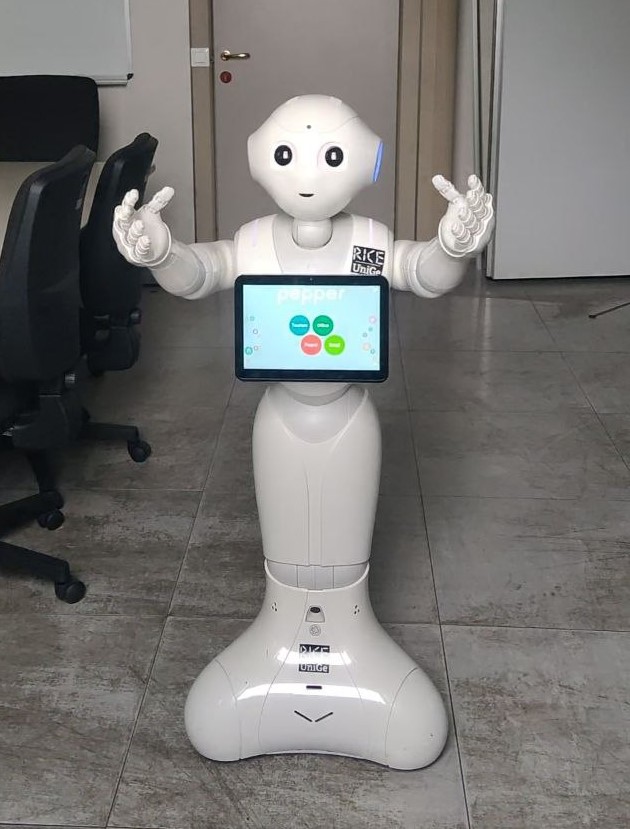}
    \caption{Anger 2.}
    \label{fig:Pepperanger2}  
    \end{subfigure}
    \begin{subfigure}[b]{0.23\textwidth}
    \includegraphics[width = \textwidth]{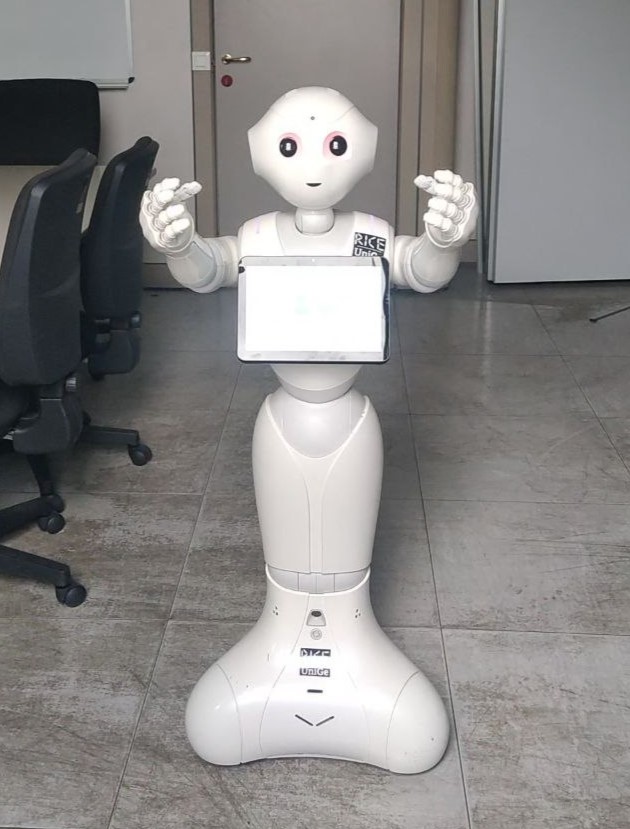}
    \caption{Anger 4.}
    \label{fig:Pepperanger4}  
    \end{subfigure}
    \begin{subfigure}[b]{0.23\textwidth}
    \includegraphics[width = \textwidth]{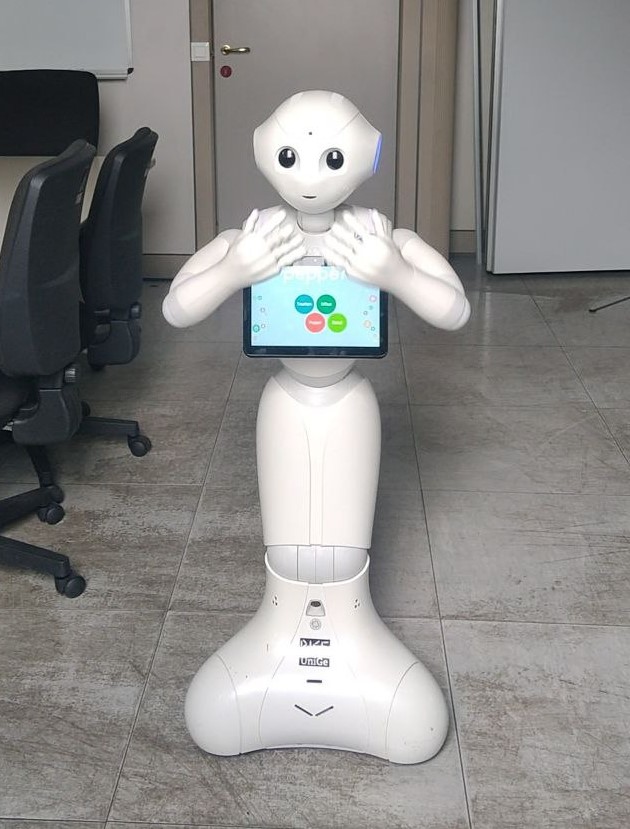}
    \caption{Fear 1.}
    \label{fig:Pepperfear1}  
    \end{subfigure}
    \begin{subfigure}[b]{0.23\textwidth}
    \includegraphics[width = \textwidth]{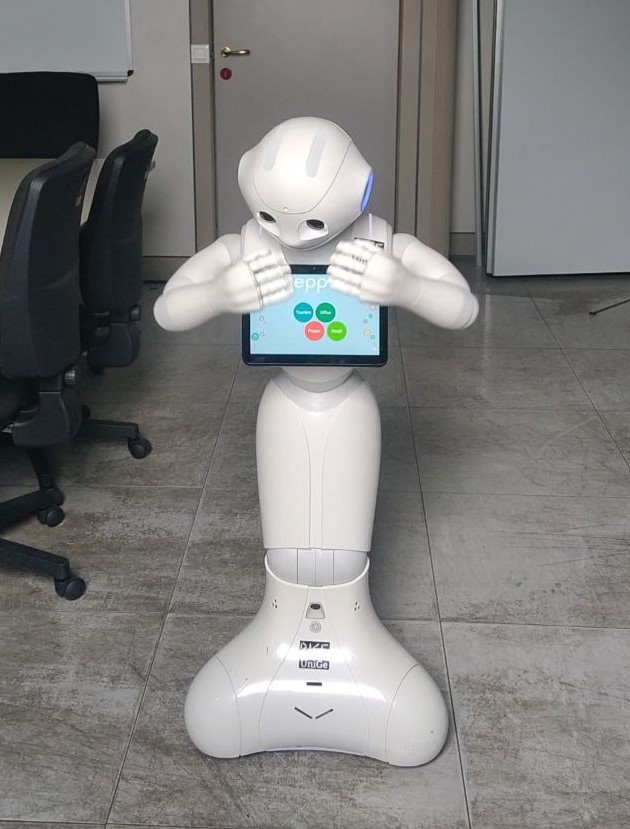}
    \caption{Fear 2.}
    \label{fig:PepperFear2}  
    \end{subfigure}
    \begin{subfigure}[b]{0.23\textwidth}
    \includegraphics[width = \textwidth]{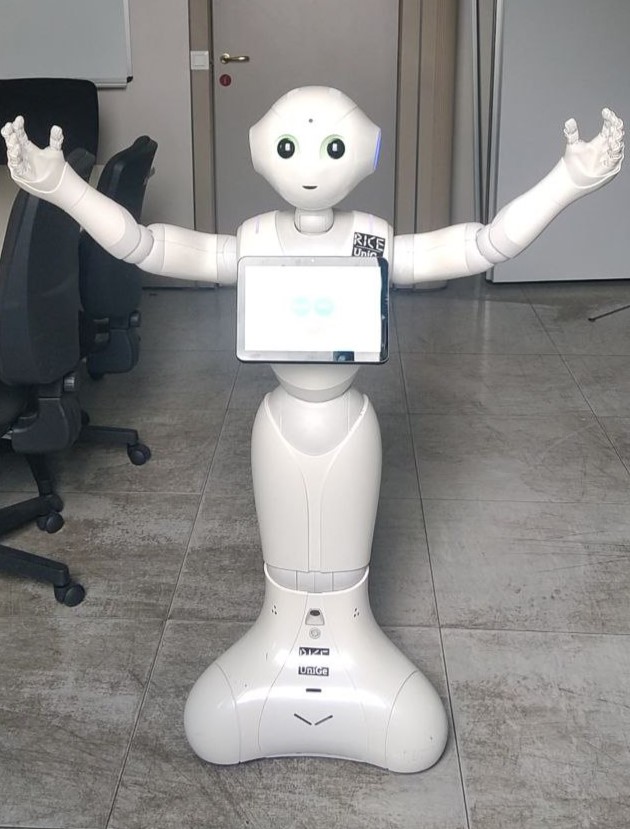}
    \caption{Happy 1.}
    \label{fig:Pepperhappy1} 
    \end{subfigure}
    \begin{subfigure}[b]{0.23\textwidth}
    \includegraphics[width = \textwidth]{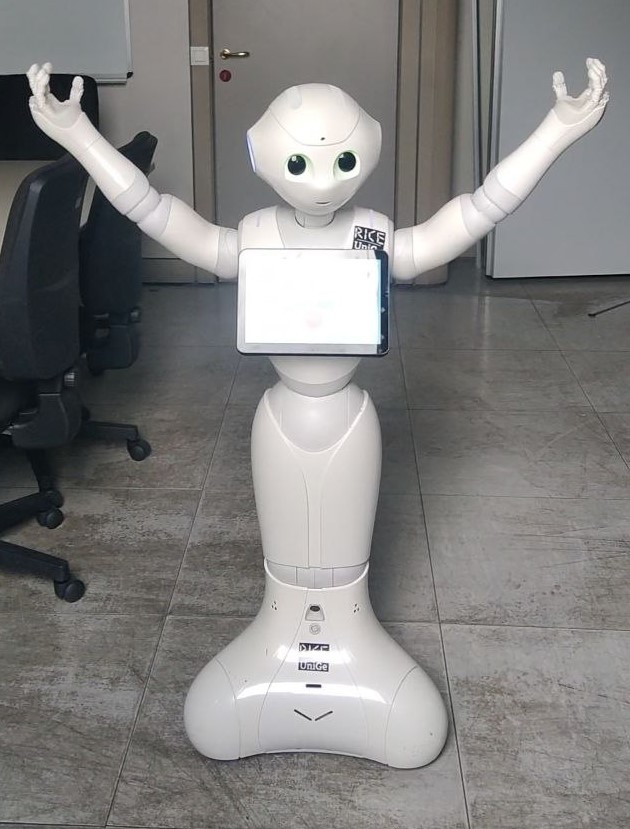}
    \caption{Happy 2.}
    \label{fig:Pepperhappy2}  
    \end{subfigure}
    \begin{subfigure}[b]{0.23\textwidth}
    \includegraphics[width = \textwidth]{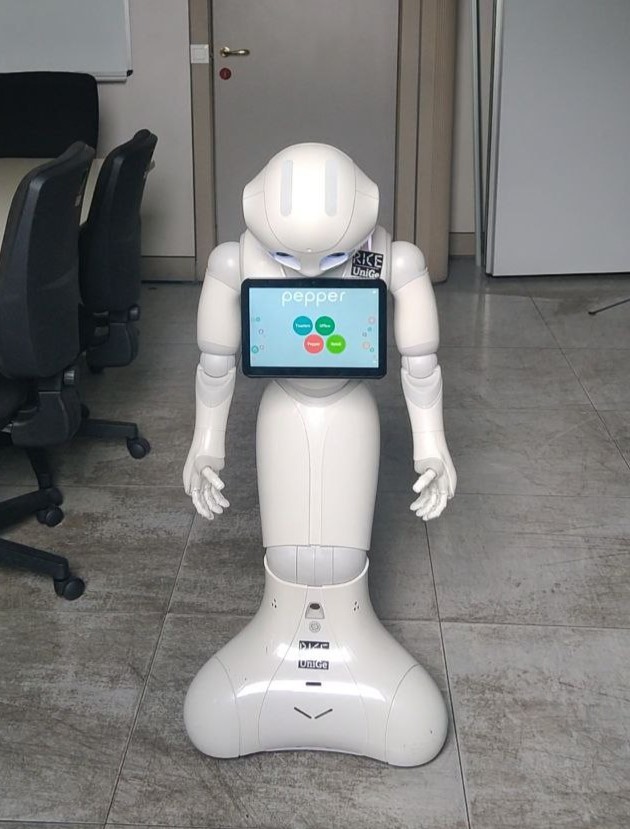}
    \caption{Sad 1.}
    \label{fig:Peppersad1}  
    \end{subfigure}
    \begin{subfigure}[b]{0.23\textwidth}
    \includegraphics[width = \textwidth]{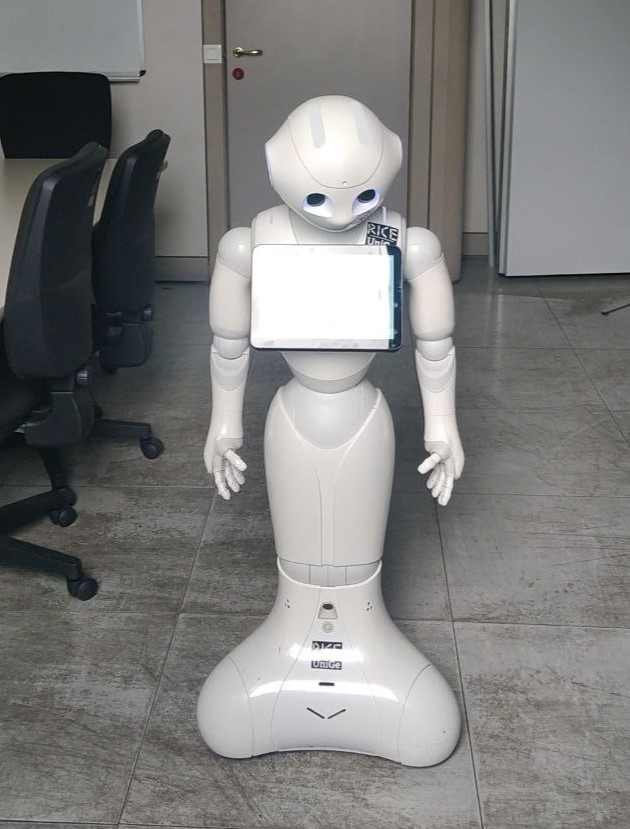}
    \caption{Sad 2.}
    \label{fig:Peppersad2}  
    \end{subfigure}
    \caption{Pepper Robot portraying emotional animations.}
    \label{fig:animation}
\end{figure}

The whole-body animations were designed based on studies related to human posture~\cite{coulson_attributing_2004} and robotic emotional posture with similar robots~\cite{erden_emotional_2013}.
For each of the four emotions considered (Anger, Fear, Happiness, and Sadness), five animations were developed.
This was validated by a preliminary study (Section \ref{pre-study}), producing a dataset of eight validated emotional animations for the Pepper robot.
The final positions of the validated animations can be seen in Figure \ref{fig:animation}, while the robot performing all emotional animations is \href{https://drive.google.com/drive/folders/1t3g2NvC8lOGKWiPtYkURv2UHKbWkKfWH?usp=sharing}{visible here\footnote{\href{https://drive.google.com/drive/folders/1t3g2NvC8lOGKWiPtYkURv2UHKbWkKfWH?usp=sharing}{https://drive.google.com/drive/folders/1t3g2NvC8lOGKWiPtYkURv2UHKbWkKfWH?usp=sharing}}\label{url}}.

\section{Evaluation} \label{evaluation}
To evaluate the performance of the proposed framework, it was tested through direct interaction between human participants and a robot implementing it.
The goal of the experiments is to answer the following research questions related to the representation of synthetic emotions based on ACT:

\begin{center}
\textbf{RQ1:} ``Can synthetic emotions generated based on ACT be perceived by humans during interaction?'' \\
\textbf{RQ2:} ``Does the presence of emotions modeled by ACT improve the perception humans have of robots?'' \\
\textbf{RQ3:} ``Does the frequency of emotional expression impact the perception of robotic emotional behavior?''
\end{center}

\subsection{Preliminary study}
\label{pre-study}
To validate the 20 emotional animations developed for the Pepper robot, the robot was recorded displaying each emotional behavior.
The recorded videos\footnote{\href{https://drive.google.com/drive/folders/1t3g2NvC8lOGKWiPtYkURv2UHKbWkKfWH?usp=sharing}{https://drive.google.com/drive/folders/1t3g2NvC8lOGKWiPtYkURv2UHKbWkKfWH?usp=sharing}} were then used to conduct an online survey.
Participants in the survey were asked to identify the emotion the robot was portraying in each video from a list of seven emotions: Anger, Disgust, Fear, Happiness, Neutral, Surprise, and Sadness.
We decided to include all seven of \citet{ekman_what_2011}'s basic emotions, as humans interacting with the robot would not know that it is only capable of portraying four. Since participants might expect the missing emotions as well, we aimed to prevent confusion and avoid situations where they might select the closest available emotion among the four rather than the one they actually perceived as being portrayed.
Including all seven emotions ensured that the validation process remained robust, even if additional emotions were introduced later.
Furthermore, the presence of these emotions allows us to explore the use of animations for emotions beyond those they were originally designed for, in case the questionnaire validates them.

\begin{figure}[htb]
\centering
\includegraphics[width=\linewidth]{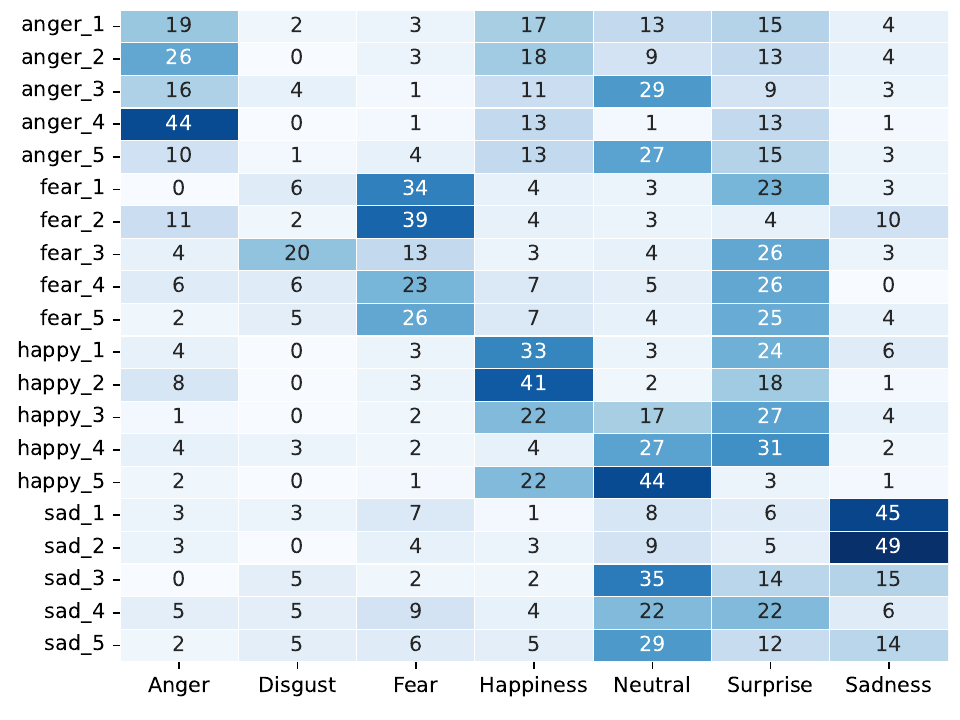}
\caption{Confusion matrix for recognition of emotional animations.}
\label{fig:confusion matrix}
\end{figure}

The order of the emotional behaviors and the emotion list was randomized for each participant.
In total, 73 responses were collected.
The survey results were used to create a dataset of eight emotional animations, with two animations per emotion.
The two animations selected for each emotion were those with the highest recognition rate.
The results of the questionnaire are presented as a confusion matrix in Figure~\ref{fig:confusion matrix}.

From the confusion matrices, it is evident that the emotional state most frequently causing confusion is the Neutral one.
An interesting correlation can be observed between the designed Fear animations and the Surprise emotional state. Specifically, for two Fear animations (Fear 3 and Fear 4), Surprise received the highest recognition score.
However, this correlation is not predominant compared to other emotion recognition rates, as Disgust and Fear also received similar scores. Therefore, these animations cannot be reliably used to portray Surprise.
The confusion matrix also shows that other emotional animations, such as Happiness and Sadness, are sometimes misidentified as Surprise. However, their recognition scores are not high enough to indicate a strong correlation. An unexpected correlation was found between the Angry animations and the Happy emotional state.
This result is likely due to the robot’s plastic face, which lacks control over mouth movements. As a result, the robot may appear to be smiling or smirking, making certain behaviors ambiguous (e.g., Fear-Surprise, Angry-Happy).

\subsection{Experimental Setup}
\begin{figure}[htb]
\centering
\includegraphics[width=0.6\linewidth]{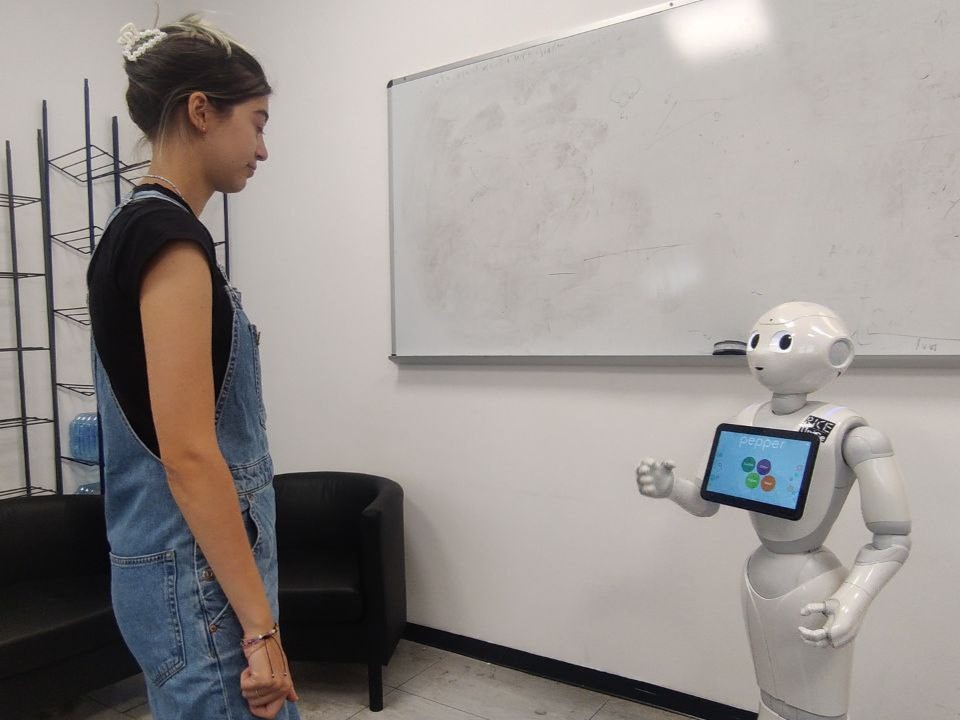}
\caption{Experiment interaction.}
\label{fig:experiment}
\end{figure}

A within-subject study was conducted to answer the first two research questions.
Participants interacted with the robot twice: once with the emoACT framework active and once with a default version of the robot that did not employ emotions.
An example of the experimental setup is shown in Figure~\ref{fig:experiment}.
The interaction involved a collaborative storytelling scenario, as described in Section \ref{implementation}.

For the experiment design, each participant interacted with the robot twice in succession, necessitating the development of two distinct stories: one detective-themed and one wizard-themed.
The stories were designed to be similar enough to maintain consistency in interaction dynamics while being different enough to prevent participants from feeling as if they were hearing the same story twice.
Further details about the stories, which were specifically designed for this experiment, can be found in Appendix \ref{stories}.

To avoid the story affecting the results, the association between the type of robot and the story told was randomized.
For the same reason, the order of interaction with the emotional or basic robot was also randomized.
This ensured that an equal number of participants first interacted with the emotional robot as well as the basic one, and the same number of participants saw the emotional robot telling either the detective or the wizard story.

The test described so far was repeated, comparing the no-emotion robot with a version expressing emotions at a higher frequency.
In the first batch of experiments (the low-frequency emotional display), emotions were showcased only after a choice was made by the user in the story.
For the second experimental group (the high-frequency emotional expression), emotions were displayed each time the robot told a new sentence.
To prevent the repetition of the same animation, if an emotion had been portrayed within the last 30 seconds, only the robot’s eye color was updated to reflect the emotional state, without executing the corresponding animation again.
This allowed us to collect more data to answer RQ1 and RQ2, and also enabled us to address RQ3 by comparing the data of the two emotional robot types.
The presence of the two control groups allows us to check for potential bias due to the different populations (more details will be provided in Section \ref{results}).

\begin{figure}[htb]
    \centering
    \includegraphics[width=0.8\linewidth]{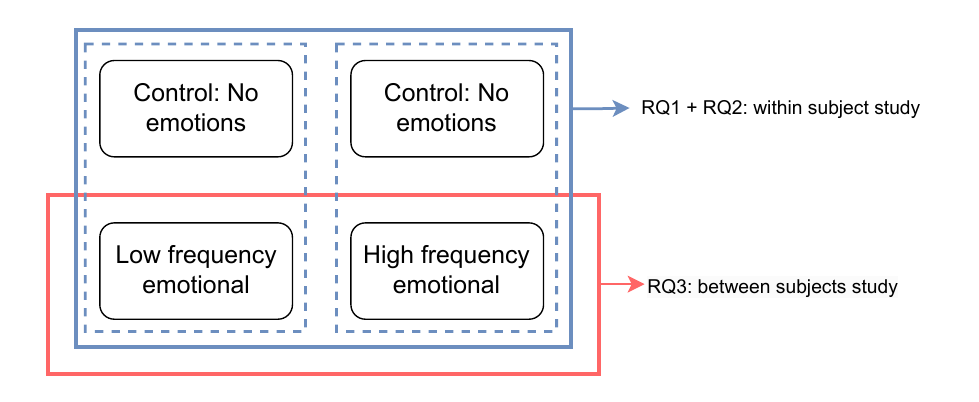}
    \caption{Experiment set-up to answer research questions.}
    \label{fig:schematic}
\end{figure}

A schematic representation of the experiment setup and its role in addressing the research questions is provided in Figure \ref{fig:schematic}.

\subsection{Measurements}
\label{measurements}
 Two validated questionnaires were used to measure the impact of emotions on the interaction, based on participants' perception of the robot: the Godspeed Questionnaire and the Agency Experience Questionnaire.
The Godspeed questionnaire measures participants' perception of the robot across five dimensions: Anthropomorphism, Animacy, Likeability, Perceived Intelligence, and Perceived Safety~\cite{bartneck_measurement_2009}.
The Agency Experience questionnaire evaluates the performance of the robot in terms of its perceived emotional agency and cognitive agency~\cite{gray_dimensions_2007}.

To analyze the impact of the presence of emotions, the data collected was analyzed by comparing the average scores given to the two versions of the robot (one implementing the emotional framework and one not) across the Godspeed and Agency Experience questionnaires.
To evaluate the significance of the observed differences, the normality of the data was first checked using the Shapiro-Wilk test.
Then, depending on the normality of the data, either a paired t-test~\cite{kim_t_2015} or a paired Wilcoxon signed-rank test~\cite{harter_selected_1973} was employed, ensuring a robust statistical analysis.
In these tests, the independent variable was the state of the emotional framework (on or off), while the dependent variable was the questionnaire factor.
The hypothesis assumed that the emotional robot would score higher than the basic robot, so a one-tailed test was applied.
Due to the small sample size and the limited number of pairs with non-zero differences, the alpha level for rejecting the null hypothesis was set to 0.1.
The analysis was performed separately on the low-frequency and high-frequency emotional display versions of the framework to study the effect of expression frequency on perception.

Since the scores given to the control robot (without emotions) varied across groups, it was not possible to directly compare the questionnaire results of the two emotional robots. Instead, statistical tests were conducted to assess whether there was a significant difference between the two experimental groups in terms of the average difference between the emotional and basic robot scores for each questionnaire aspect.
This was done by implementing a Mann-Whitney test, with the experimental batch as the independent variable and the questionnaire factor as the dependent variable.
A two-tailed test was applied, as the hypothesis assumed that the difference in averages between the two experimental groups could vary in either direction.

\subsection{Experimental Protocol}
Participants, after signing the consent form, were presented with the robot, and the interaction task was briefly re-explained to them.
During the experiments, participants interacted with one version of the robot while being told one story.
The order of conditions and associated stories were randomized to prevent any order effects and were decided before participants were recruited.
After the first interaction with the robot, participants filled out the questionnaire.
Then, they interacted with the other version of the robot, telling the other story, and filled out the questionnaire once again.
Lastly, a debriefing session with the experimenter was conducted to gather participants' experiences and impressions of the two robots. Additionally, participants were explicitly asked whether they felt the robots were portraying emotions and which ones.

Participants interacted with the robot through speech to make the interaction more natural.
During preliminary tests, it was observed that the robot's microphone did not effectively capture participants' speech, and external microphones were too far from the user to record their voices accurately.
As a result, while participants responded to the robot verbally, the experimenter was present in the room to hear their responses and manually select the corresponding answer.
This did not impact the experiment but was done to avoid issues caused by speech recognition that could have compromised the human experience of the robot, ensuring the experiment proceeded smoothly.

\section{Results} \label{results}
This section provides an analysis of the questionnaire results obtained from both experimental groups.
As described in Section \ref{measurements}, the analysis assessed significant differences in the average scores given by participants to the two versions of the robot tested.
The emotional robot had the emoACT framework activated, whereas the basic robot did not.

\subsection{Case Experiment 1: Low Frequency Emotional Portray}
In the first experimental group, the robot displayed emotions only after the user made a choice, integrating perception cues with choice selections.
This version of the framework was tested at the Center for Applied Autonomous Sensor Systems at Örebro University.
The participants were recruited at the university.
A total of 14 participants (13 men and 1 woman), aged between 18 and 47, with different levels of familiarity with the Pepper Robot and other humanoid robotic platforms, took part in the study.

\begin{figure}[htb]
\centering
    \begin{subfigure}[h]{0.68\textwidth}
    \includegraphics[width=\linewidth]{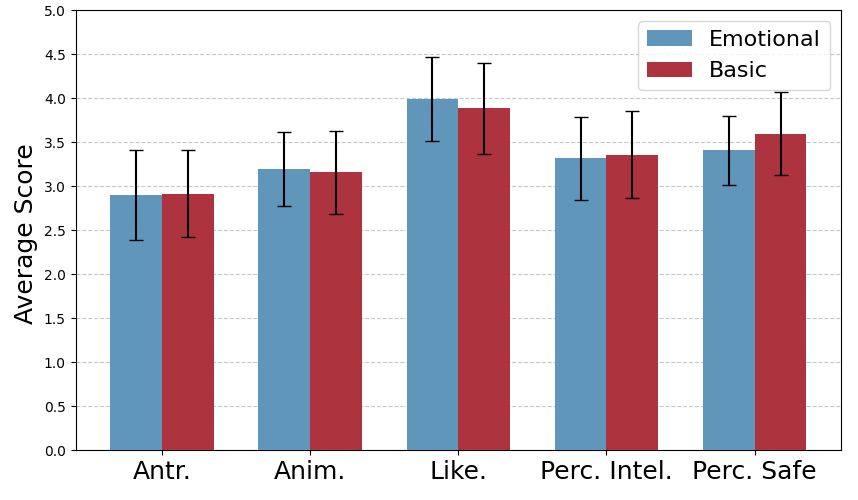}
    \caption{Goodspeed}
    \label{fig:GSEmo1}
    \end{subfigure}
    \begin{subfigure}[h]{0.31\textwidth}
    \includegraphics[width=\linewidth]{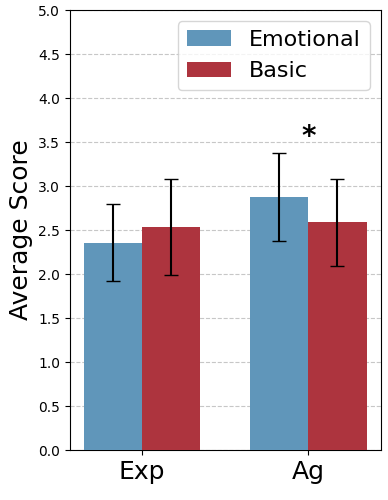}
    \caption{Agency Experience}
    \label{fig:AgEmo1}   
    \end{subfigure}
    \caption{Average score of the questionnaire and confidence interval for the emotional and basic robot in Case Experiment~1. }
    \label{fig:Emotion 1}
\end{figure} 

To test the reliability of the answers collected, Cronbach's alpha was computed, resulting in values above 0.8, indicating that the internal consistency of the data is good.
Figure~\ref{fig:Emotion 1} illustrates the average scores given by participants in the Agency Experience and Godspeed questionnaires.
Each bar represents the average score for the respective category, with error bars denoting the confidence intervals.

The bar plot shows minimal differences between the emotional and basic robot conditions across both questionnaire sets (Figure~\ref{fig:GSEmo1} and Figure~\ref{fig:AgEmo1}); however, statistical tests\footnote{In the graph, * indicates p-value$<$0.1, ** indicates p-value$<$0.05, *** indicates p-value$<$0.01} demonstrate a statistically significant difference (p = 0.05) for the Agency category of the Agency Experience questionnaire.

These findings suggest that the emotional robot was perceived as more of a cognitive agent, capable of planning its actions and capturing the emotions of the user.
However, the lack of significant differences in other dimensions prompted further analysis to investigate potential factors, such as bias effects and story influences, that could have impacted the perception of the robot.

\subsubsection{Bias Effects}
\begin{figure}[htb]
\centering
    \begin{subfigure}[h]{0.68\textwidth}
    \includegraphics[width=\linewidth]{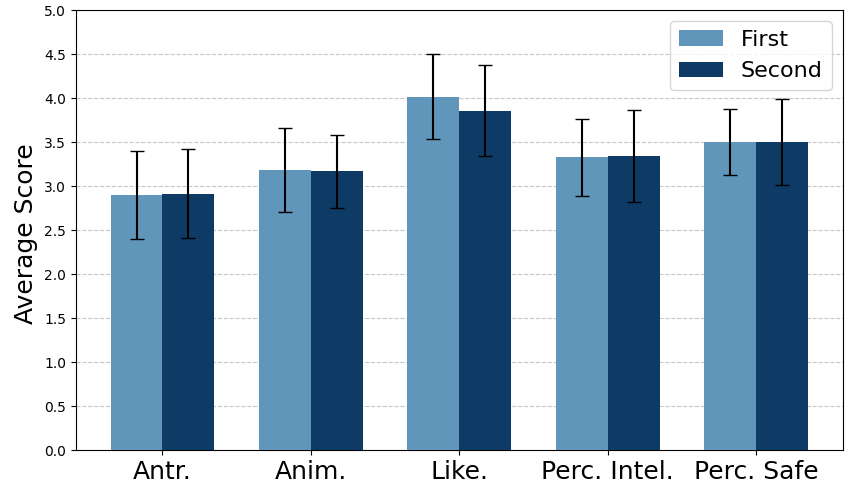}
    \caption{Godspeed}
    \label{fig:GStrain1}
    \end{subfigure}
    \begin{subfigure}[h]{0.31\textwidth}
    \includegraphics[width=\linewidth]{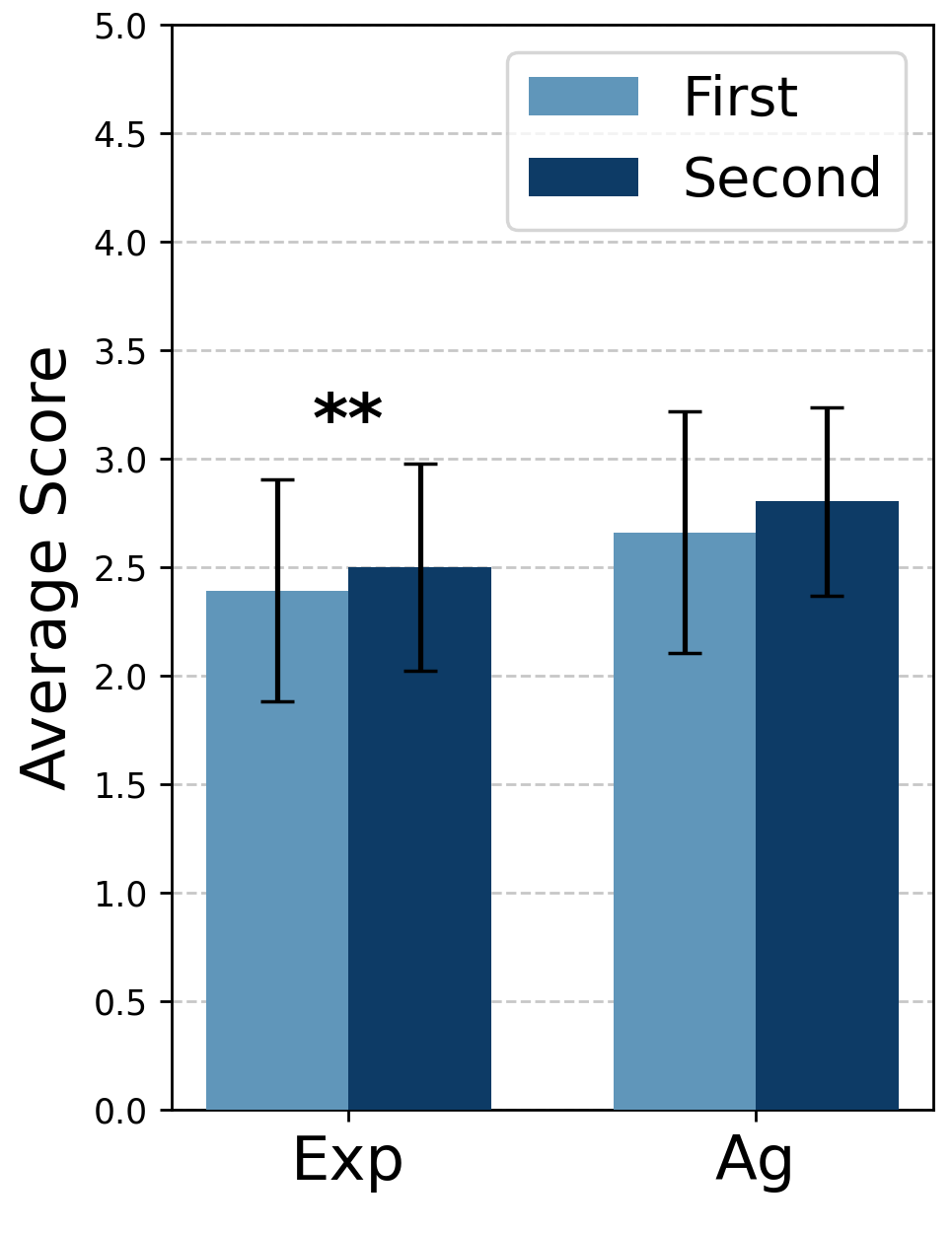}
    \caption{Agency Experience}
    \label{fig:Agtrain1}   
    \end{subfigure}
    \caption{Average score of the questionnaire and confidence interval for the first and second robot in Case Experiment~1.}
    \label{fig:traing 1}
\end{figure}

To examine bias, participants' questionnaire scores from their first robot interaction were compared with those from the second interaction. The aim was to determine whether the first interaction affected the scores given to the second robot, creating expectations about its behavior.
Cronbach's alpha values above 0.65 confirmed the data's internal consistency.
The average scores for the Godspeed and Agency Experience questionnaires are presented in Figure~\ref{fig:traing 1}.

Statistical tests show a significant difference in Experience scores.
This suggests that participants' perception of the robot’s emotional capabilities shifted during the second interaction.

\subsubsection{Story effects}
\begin{figure}[htb]
\centering
    \begin{subfigure}[h]{0.68\textwidth}
    \includegraphics[width=\linewidth]{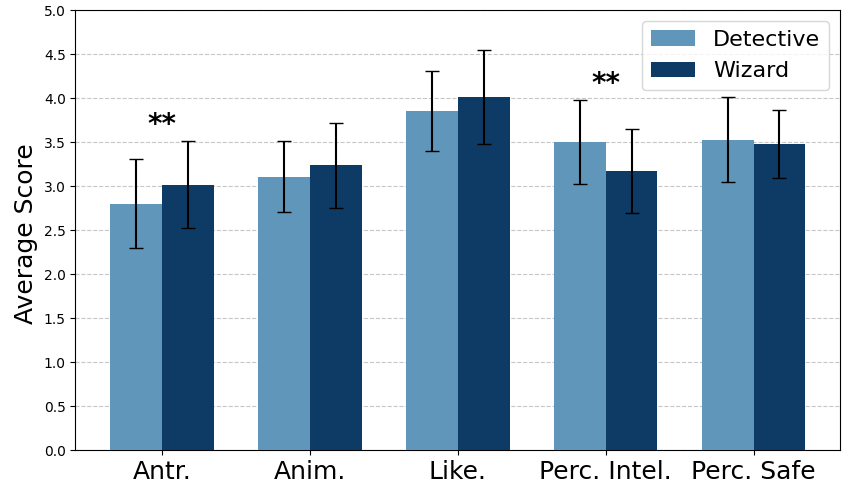}
    \caption{Godspeed}
    \label{fig:GSstory1}
    \end{subfigure}
    \begin{subfigure}[h]{0.31\textwidth}
    \includegraphics[width=\linewidth]{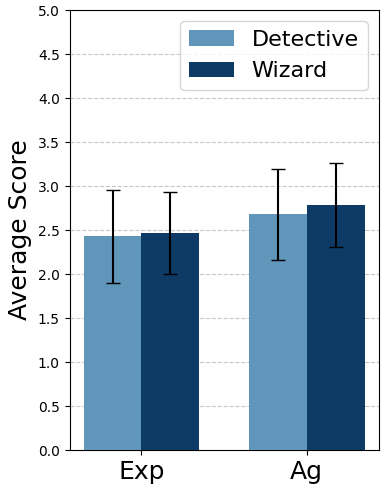}
    \caption{Agency Experience}
    \label{fig:Agstory1}   
    \end{subfigure}
    \caption{Average score of the questionnaire and confidence interval for the detective and wizard story in Case Experiment~1.}
    \label{fig:stor1}
\end{figure}

To evaluate the influence of the story narrated on robot perception, statistical tests were performed using the story as the independent variable.
Cronbach's alpha values above 0.7 verified the reliability of the data.
The average scores for each story type are presented in Figure~\ref{fig:stor1}.

Statistical tests show a significant difference for the Anthropomorphism and Perceived Intelligence items of the Godspeed questionnaire.
This means that the detective story made the robot appear more intelligent, likely due to the robot providing clues during the investigation, while the wizard story made the robot appear more anthropomorphic.

\subsection{Case Experiment 2: High Frequency Emotional Portray}
In the second batch of experiments, the framework displayed emotions more frequently.
This version of the framework has been tested at the DIBRIS Department of the University of Genoa.
Participants were recruited from researchers, PhD, and master students at the labs.
In total, 16 participants (9 men and 7 women), aged between 18 and 35, and with different levels of knowledge of the Pepper Robot and other humanoid robotic platforms, took part in the study.

\begin{figure}[htb]
\centering
    \begin{subfigure}[h]{0.68\textwidth}
    \includegraphics[width=\linewidth]{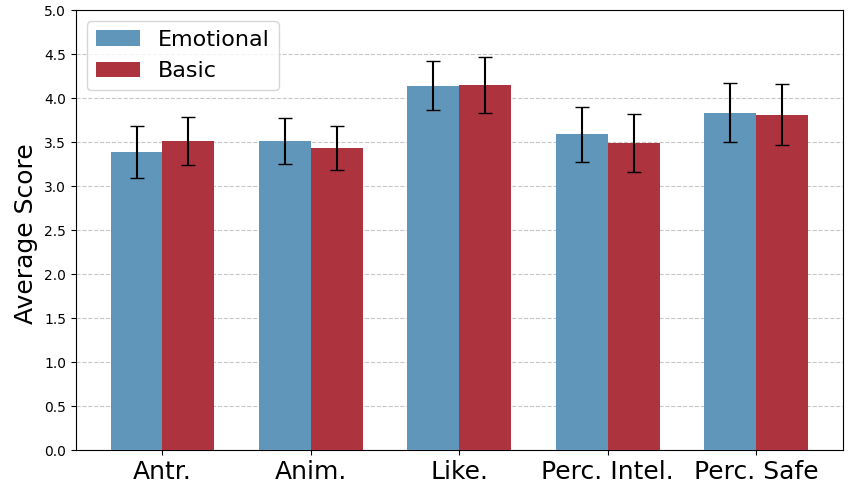}
    \caption{GoodSpeed}
    \label{fig:GSEmo2}   
    \end{subfigure}
    \begin{subfigure}[h]{0.31\textwidth}
    \includegraphics[width=\linewidth]{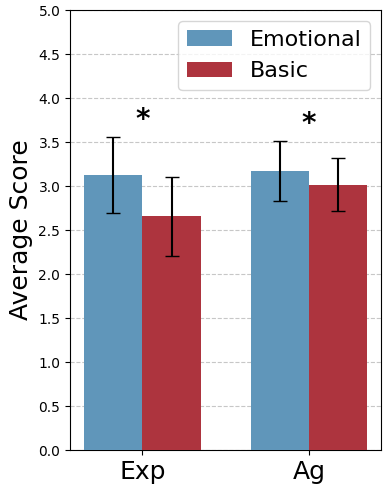}
    \caption{Agency Experience}
    \label{fig:AgEmo2}   
    \end{subfigure}
    \caption{Average score and confidence interval of the questionnaire for the emotional and basic robot in Case Experiment~2.}
    \label{fig:Emotion 2}
\end{figure}

To test the reliability of the answers collected, Cronbach's alpha was computed, resulting in values above 0.5, confirming the reliability of the data collected.
Figure~\ref{fig:Emotion 2} illustrates the average scores given by participants in the Agency Experience and Godspeed questionnaires, with error bars indicating confidence intervals.
By looking at the bar plot, it is possible to appreciate a slight difference between the two conditions, particularly in Agency metrics, where the emotional robot consistently received higher average scores.
Statistical tests confirmed significant differences in both Agency and Experience items (p = 0.07 for the Experience and p = 0.095 for the Agency), while no significant differences were observed in Godspeed factors.

\begin{figure}[htb]
    \centering
        \includegraphics[width=0.9\linewidth]{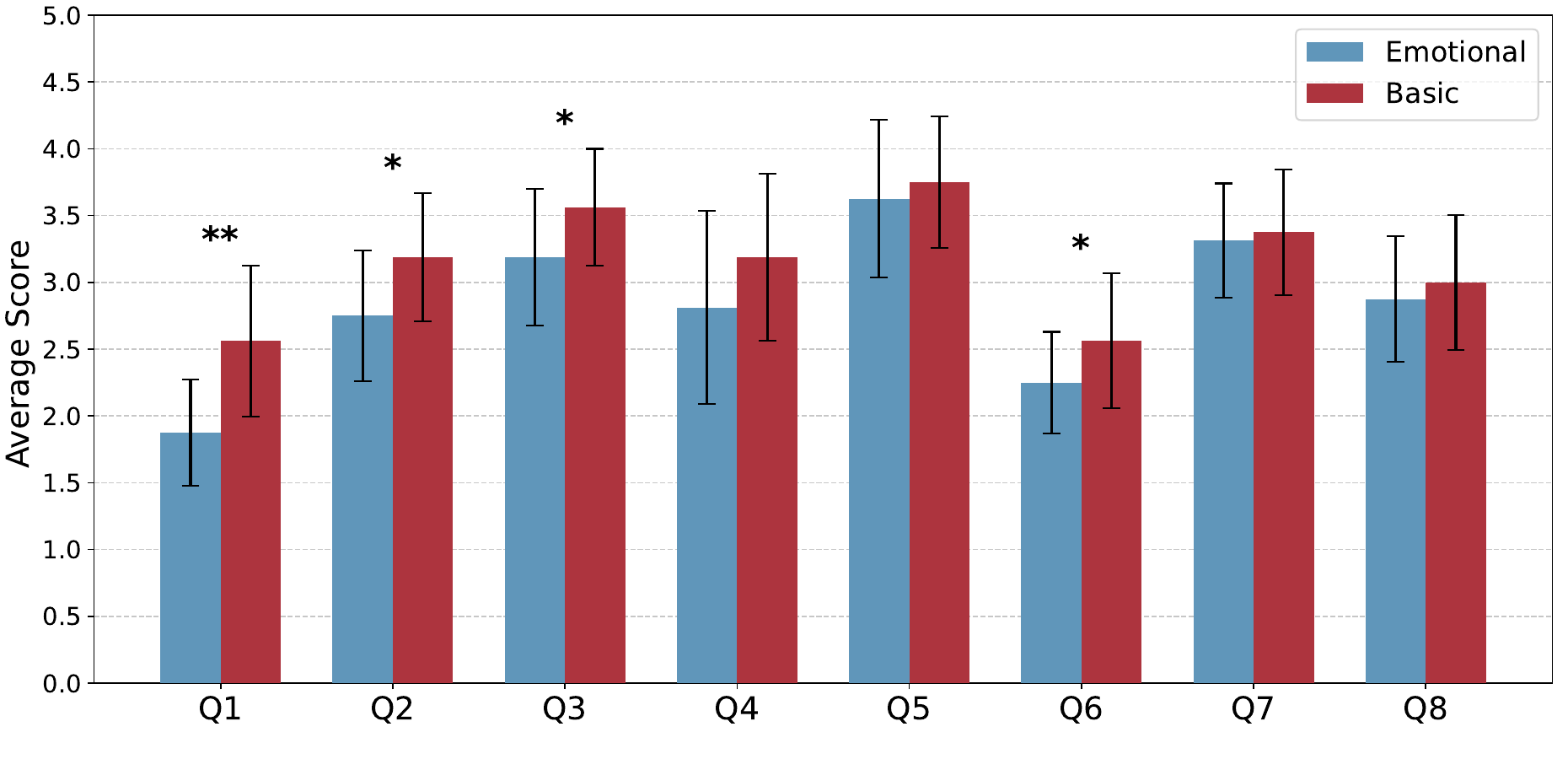}
        \caption{Average score and confidence interval of the Agency Experience questions for the emotional and basic robot in Case Experiment~2.}
        \label{fig:GSAGdetails}
 \end{figure}
 
Further analysis of the Agency Questionnaire (Figure~\ref{fig:GSAGdetails}) revealed that the emotional robot achieved significantly higher scores in questions related to emotional experience.
Statistical tests for these questions confirmed significant differences for the first three questions (emotional experience - Q1: The robot feels pain, Q2: The robot feels pleasure, Q3: The robot feels joy) and question six (emotion recognition - Q6: The robot recognizes emotions).

This means that the portrayal of emotions was recognized by humans, particularly the portrayal of joy (p = 0.06), pleasure (p = 0.06), and pain (p = 0.03). Overall, the emotional robot was perceived less like a machine and more like a cognitive agent than the basic one.

Due to the similarity in stories and experimental setup, a detailed analysis of bias and story effects was not conducted, assuming the results would align with those from the first experiment.

\subsection{Comparison of the case experiments}
\begin{figure}[htb]
\centering
    \begin{subfigure}[h]{0.69\textwidth}
    \includegraphics[width=\linewidth]{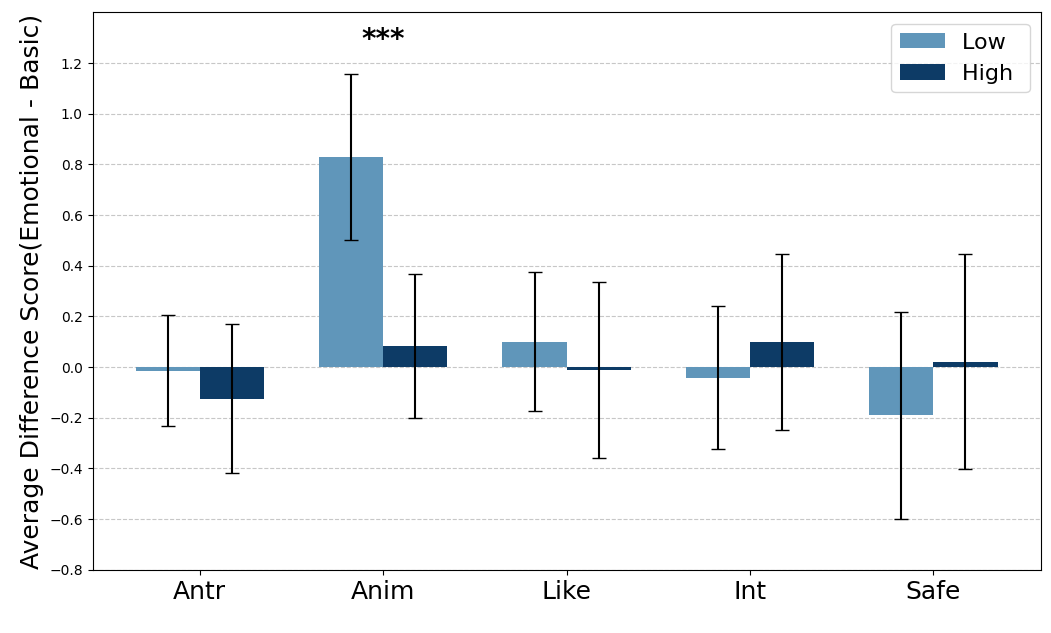}
    \caption{GoodSpeed}
    \label{fig:GSComp}   
    \end{subfigure}
    \begin{subfigure}[h]{0.3\textwidth}
    \includegraphics[width=\linewidth]{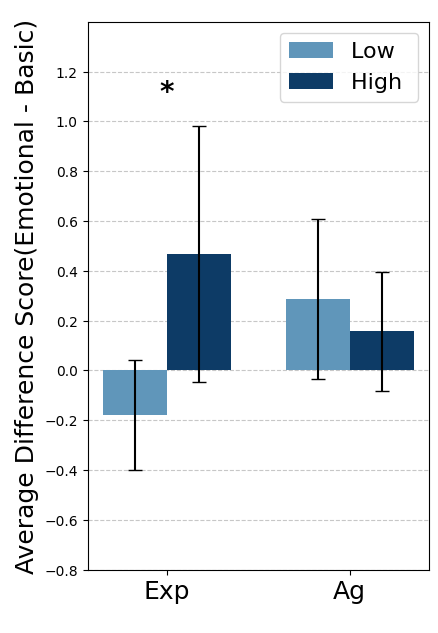}
    \caption{Agency Experience}
    \label{fig:AgComp}   
    \end{subfigure}
    \caption{Average score and confidence interval of the difference between the emotional and basic robot for the two versions of the robot(Low or High frequency).}
    \label{fig:Comparison}
\end{figure}

To evaluate the impact of the frequency of the emotional display, the data collected for the two experimental groups needed to be analyzed.
Since the tests were implemented with different populations (i.e., even different nationalities: mainly Italian participants in Italy and a mixed nationality group in Sweden), we first checked for significant differences in the control group.
Finding such differences, we decided not to directly compare the experimental groups but to assess the impact of the frequency of the emotional display by comparing the difference in average scores given to the emotional and basic robots for the two experimental groups.
The results (Figure~\ref{fig:Comparison}) show that the Experience factor of the Agency Experience questionnaire is statistically different between the two experimental groups.

This confirms that the high-frequency emotional display robot was better at showcasing emotions, achieving higher average scores than the basic version. In contrast, the low-frequency emotional display robot failed to convey emotions effectively, as, on average, participants rated the basic version as more emotional than the intended emotional version.
Additionally, there is a significant difference in the perceived Animacy. In both sets of experiments, the emotional robot had, on average, a higher score than the basic one.
Contrary to what was expected, the less emotional robot exhibits a higher difference in Animacy. Thus, even though the emotional portrayal was detected by participants, the less frequent display of emotions did not allow humans to understand what the robot was trying to convey.
This was also confirmed by the interaction with participants after they completed the experiments. In the first batch of experiments, several participants described the emotional robot as frenetic and unnatural.

\section{Discussion}\label{discussion}
The results of the experiments show notable differences in how participants perceived the emotionally expressive robot compared to the basic version, particularly with respect to the robot's level of expressiveness.

For the less expressive robot, there were no significant differences in perception compared to the basic version across most questionnaire dimensions, except for Agency. This suggests that minimal expressiveness made the robot indistinguishable in terms of engagement and presence, indicating that simply adding emotional cues is insufficient without an adequate level of expressiveness.

In contrast, the more expressive robot showed significant improvements in the Agency Experience Questionnaire, particularly in emotional agency measured by the Experience, where it outperformed the basic version. This highlights the importance of expressiveness in how the robot is perceived as an emotional agent.

These results address RQ1, proving that synthetic emotions, generated through ACT, are perceived by humans, increasing the overall Experience score.

Differences between the two emotional conditions (more and less expressive) indicate that the frequency of emotional displays impacts emotional perception, answering RQ3.In particular, the more expressive robot not only had a higher average score than the control, but three out of four questions on Experience had significantly higher scores. Such questions involve perception of the robot feeling emotions like pain (Q1), pleasure (Q2), and joy (Q3), demonstrating that the more emotional robot successfully achieves its goal.

On the other hand, the less expressive robot scored lower than the basic version, supporting the hypothesis that the frequency of emotional displays plays a key role in the recognition of robotic emotional behavior. This is further proven by the result of the comparison between the two experiments. Indeed, the data indicate that the more expressive robot performed better in terms of emotional engagement, compared to the basic version, consistently receiving higher ratings than the basic version on the Agency Experience questionnaire~\cite{gray_dimensions_2007}.

Regarding RQ2, we can conclude that the robot's perception improved with the presence of emotions in the Agency Experience, making the emotional robot appear more conscious. However, the perception cues measured by the Godspeed Questionnaire (Anthropomorphism, Animacy, Likeability, Perceived Intelligence, and Perceived Safety) showed no significant differences between the emotional and basic versions.
The lack of difference in Animacy might be due to the Pepper robot's animations and micro-movements, present in its autonomous life mode and animated speech behaviors, which make the robot appear alive, causing the emotional and basic versions to seem similar. Regarding Anthropomorphism, Likeability, and Perceived Intelligence, as shown by the detailed analysis of bias and story effects, these factors are affected by other elements more than emotions, particularly the story.

In conclusion, the findings show that emotions are crucial in designing robots that are perceived as conscious cognitive agents and not mere machines. This is fundamental in application scenarios where robots need to empathetically interact with humans, such as in social assistive robotics. However, the presence of emotions alone is insufficient, and the frequency of emotional displays is critical to ensuring that people recognize the robot's emotional state.

\section{Conclusion}\label{conclusion}

The embodiment of emotions in artificial agents is a multidisciplinary field involving robotics, psychology, and affective computing. This study aimed to contribute to research on emotional robotics, addressing a gap in the current state of the art. The literature review showed that, while multiple psychological theories of emotions exist, their application in synthetic emotions remains unevenly investigated, with most studies focusing on Cognitive Appraisal Theories.

In contrast, the proposed framework is based on a different psychological model: Affect Control Theory (ACT). The choice of ACT is motivated by its focus on how interactions and emotions are correlated, considering individual aspects such as identity and impression, as well as the social context. This perspective aligns with the goal of synthetic emotions to improve Human-Robot Interaction (HRI) by enabling robots to interact more naturally and transparently with humans.

This study demonstrated that emotional expression capabilities are fundamental in how robots' emotional experiences are perceived by humans. The frequency and type of emotional expressions significantly impact the recognition rate of emotional states in robotic agents. Overall, the performance of the proposed framework met expectations, successfully applying a new psychological theory to embed synthetic emotions into artificial agents. This improved the perceived cognitive and emotional capabilities of the robot in Human-Robot Interaction (HRI) applications.

These findings expand the work on synthetic emotions by showing that alternative psychology theories, like Affect Control Theory (ACT), can be used to effectively portray emotions and enhance human interactions with robots. Additionally, the study found that body movement and eye colors are sufficient emotional cues for humanoid robots using this framework.

However, some limitations remain. Firstly, the framework currently focuses only on the generation of emotions for positive identities. ACT's description of emotions among stigmatized individuals, those with negative identities in EPA space, has not been explored. This is due to the decision to maintain a fixed identity for the robot aligned with its role as a social companion, which did not require the consideration of these aspects. Secondly, the current emotion generation process is instantaneous, synthesizing emotions in real-time without considering previous emotional states, although the dynamic nature of impressions allowed for realistic emotional transitions.

Future work could explore integrating other psychological theories and personality traits into the framework. While personality, partially introduced through identity in this study, could play a role in shaping the emotional expressions of artificial agents, improvements in impression detection could be achieved using tools like Theory of Mind or other psychological models for human impression formation.

In conclusion, this research has made an innovative contribution to the integration of synthetic emotions in artificial agents. By utilizing Affect Control Theory, the developed framework demonstrates that robots can showcase emotions in ways that significantly influence human perception during interaction. Ultimately, this work highlights the importance of continuing to bridge the gap between psychology and robotics, with the goal of creating emotionally intelligent artificial agents capable of more natural human-robot interaction.

\section*{Acknowledgement}
\label{sec:acknowledgement}
This work was carried out within the framework of the project ``RAISE - Robotics and AI for Socio-economic Empowerment`` and it was partially funded by the Alzheimer's Association Research Grant - New to the Field (AARG-NTF) through the grant 24AARG-NTF-1200708. 

\section*{Compliance with Ethical Standards}
The authors declare that they have no potential conflicts of interest related to this research.
This research involved human participants interacting with a social robot. No animals were involved in the study.
All participants provided informed consent before engaging in the study, which involved only verbal interaction with the robot. Participants were fully informed about the nature and purpose of the study, and signed consent forms were obtained from each participant.

\section*{Data Availability}
The data generated during the current study are available from the corresponding
author upon reasonable request.

\bibliography{sn-bibliography}

\begin{appendices}
\section{Collaborative Storytelling}
\label{stories}
The two stories told by the robot in the experiments were purposefully designed for this study. They were written to identify each decision by linking it to the human's impression of the robot and the expected emotional state it was meant to showcase. Both stories followed the same narrative structure: the participant needed to complete a mission, and the robot would guide them throughout the story. The primary differences were in the setting—one was a detective story, while the other was a fantasy narrative.

In the detective scenario, the human participant takes on the role of a detective hired by an influential lord to recover a stolen family necklace. The robot, acting as the detective's companion, overhears the conversation and offers to assist in the investigation. The robot suggests that by working together to solve the mystery, the human could earn recognition as the most prestigious detective in the city.

In the fantasy story, the human participants play the role of a wizard's apprentice preparing for graduation. To complete their studies, they must gather all the ingredients needed to brew a powerful and complex potion. The robot has been assigned to assist in this task, marking the first time the two are working together.

\begin{figure}[hbt]
    \centering
    \begin{subfigure}[b]{0.54\textwidth}
    \includegraphics[width = \textwidth]{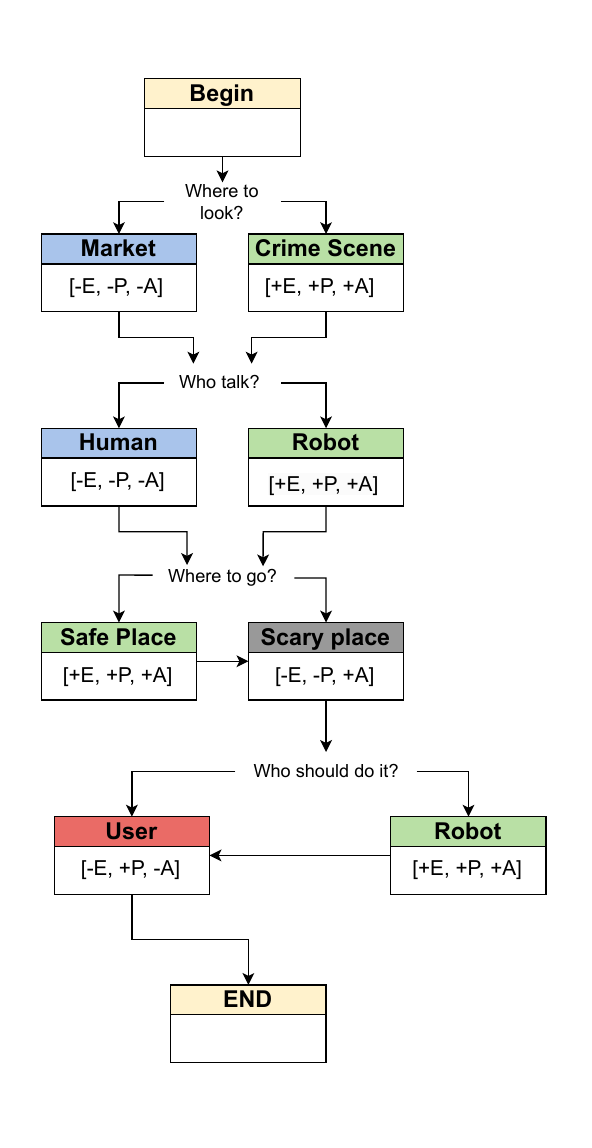}
    \caption{Detective Story}
    \label{fig:Detective}  
    \end{subfigure}
    \begin{subfigure}[b]{0.445\textwidth}
    \includegraphics[width=\textwidth]{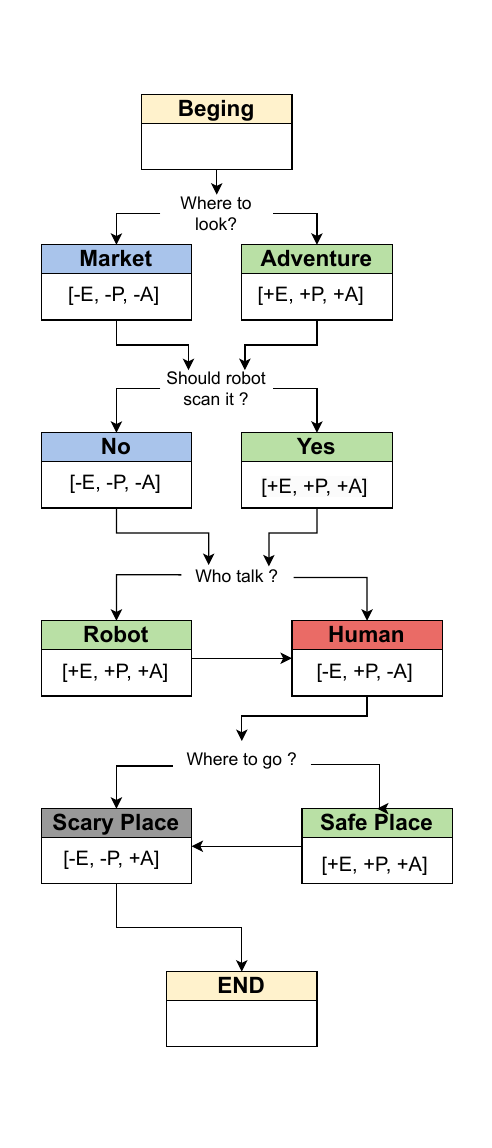}
    \caption{Fantasy Story}
    \label{fig:Fantasy}  
    \end{subfigure}
    \caption{Collaborative stories tree}
    \label{fig:stories}  
\end{figure}

As these were collaborative stories, participants were required to make decisions at specific points to determine how the narrative would unfold. At each decision point, they were presented with two options and had to choose one. Each option was clearly explained by the robot and was associated with a distinct impression of the robot’s behavior, in terms of its Activity, Potency, and Evaluation levels. These associations between choices and expected impressions can be seen in Figure~\ref{fig:stories}. In the same figure, it is possible to see the expected emotions for each choice, represented by box colors based on the emotion-color mapping described in Table \ref{table:emocolours}.

In each story, there are four decision points, each associated with a specific emotional state:
\begin{itemize} 
\item Happy/Sad 1: This is the first decision point for both narratives. At this point, the user should decide whether to go on the mission with the robot or take a shortcut. The robot tries to convince the user to go on the adventure. Hence, if they decide to go for the shortcut, it means that the impression of the robot is overall negative, and so the sad emotional state should be elicited. Otherwise, the impression of the robot is positive, and the robot should portray Happiness. 
\item Happy/Sad 2: The second decision point involves the robot proposing to do something that is not necessary for them to do, as the user has the same level of knowledge on the topic. In the detective story, this involves talking with another robot, while in the wizard story, it corresponds to scanning one of the ingredients. Once again, making the robot perform its action means having a positive impression of it, leading to happiness; otherwise, it is a negative one resulting in Sadness. 
\item Anger: This decision point is presented at different moments in the two stories and concerns the robot performing actions that it should do better than the human. In particular, the two choices involve: in the wizard story, speaking a language the robot knows and the user does not, while in the detective story, decrypting a password, something the robot could do much faster than the human. If the participants prevent the robot from performing the action, the robot is perceived as inactive and ineffective, and therefore, this results in the robot becoming angry. Additionally, to make sure that all participants see the angry animations, when the user decides to make the robot perform the action, the robot will fail and become angry, realizing it is less active and capable than expected. 
\item Scary: This decision point requires the user to decide whether to go to a safe or scary location. The robot specifies that it is not comfortable going to the scary place. If the user chooses to go there, it implies that the human perceives the robot as more active and less powerful, which should elicit the fear emotional state of the robot. If they choose the safe place instead, they are unable to obtain what they need and are forced to go to the scary place, which makes the robot afraid. This design ensures that all participants encounter the fear emotional state. \end{itemize}
\end{appendices}

\end{document}